\documentclass[authoryear,3p,times,twocolumn]{elsarticle}
\usepackage{amsthm}
\usepackage{footnote}
\usepackage{amsmath}
\usepackage{algpseudocode}
\usepackage[ruled, vlined, linesnumbered, resetcount]{algorithm2e}
\usepackage{subfig}
\usepackage{placeins}
\usepackage[normalem]{ulem}
\usepackage{enumitem}
\usepackage{multirow}
\useunder{\uline}{\ul}{}

\usepackage{amssymb}
\usepackage{graphicx}

\usepackage{subfig}
\usepackage{multirow}

\journal{}

\begin{document}

\begin{frontmatter}



\title{Data Driven Exploratory Attacks on Black Box Classifiers in Adversarial Domains}


\author[tj]{Tegjyot Singh Sethi \corref{cor1}}
\ead{tegjyotsingh.sethi@louisville.edu}
\author[tj]{Mehmed Kantardzic}
\ead{mehmedkantardzic@louisville.edu}


\address[tj]{Data Mining Lab, University of Louisville, Louisville, USA }
\cortext[cor1]{Corresponding author.}

\begin{abstract}

While modern day web applications aim to create impact at the civilization level, they have become vulnerable to adversarial activity, where the next cyber-attack can take any shape and can originate from anywhere. The increasing scale and sophistication of attacks, has prompted the need for a data driven solution, with machine learning forming the core of many cybersecurity systems. Machine learning was not designed with security in mind, and the essential assumption of stationarity, requiring that the training and testing data follow similar distributions, is violated in an adversarial domain. In this paper, an adversary's view point of a classification based system, is presented.  Based on a formal adversarial model, the \textit{Seed-Explore-Exploit} framework is presented, for simulating the generation of data driven and reverse engineering attacks on classifiers. Experimental evaluation, on 10 real world datasets and using the Google Cloud Prediction Platform, demonstrates the innate vulnerability of classifiers and the ease with which evasion can be carried out, without any explicit information about the classifier type, the training data or the application domain. The proposed framework, algorithms and empirical evaluation, serve as a white hat analysis of the vulnerabilities, and aim to foster the development of secure machine learning frameworks.

\end{abstract}


\end{frontmatter}


\section{Introduction}
\label{sec:introduction}

The growing scale and reach of modern day web applications has increased its reliance on machine learning techniques, for providing security. Conventional security mechanisms of firewalls and rule-based black and white lists, cannot effectively thwart evolving attacks at a large scale \citep{saha2014application
}. As such, the use of data driven machine learning techniques in cybersecurity applications, has found widespread acceptance and success \citep{
big2013big}. Whether it be for outlier detection for network intrusion analysis \citep{zamani2013machine}, biometric authentication using supervised classification \citep{d2014avatar}, or for unsupervised clustering of fraudulent clicks \citep{walgampaya2011cracking}, the use of machine learning in cybersecurity domains is ubiquitous. However, during this era of increased reliance on machine learning models, the vulnerabilities of the learning process itself have mostly been overlooked. Machine learning operates under the assumption of stationarity, i.e. the training and the testing distributions are assumed to be identically and independently distributed (IID) \citep{vzliobaite2010learning}. This assumption is often violated in an adversarial setting,  as adversaries gain nothing by generating samples which are blocked by a defender's system \citep{guerra2010exploring}. The dynamic and contentious nature of this domain, demands a thorough analysis of the dependability and security of machine learning systems, when used in cybersecurity applications. 

\begin{figure}[t]
\centering
\subfloat[An adversary making probes to the black box model \textit{C}, can learn it as \textit{C}', using active learning.]{\includegraphics[width=1\linewidth]{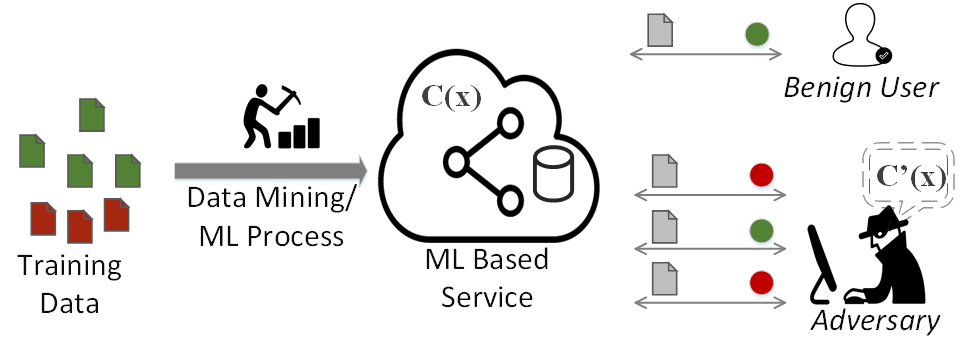}} \\

\subfloat[Example task of attacking behavioral CAPTCHA. Black box model $C$, based on Mouse Speed and Click Time features, is used to detect benign users from bots. Adversary can reverse engineer \textit{C} as $C'$, by guessing click time feature and making probes based on the human response time chart, using the same input channels as regular users.]{\includegraphics[width=1\linewidth]{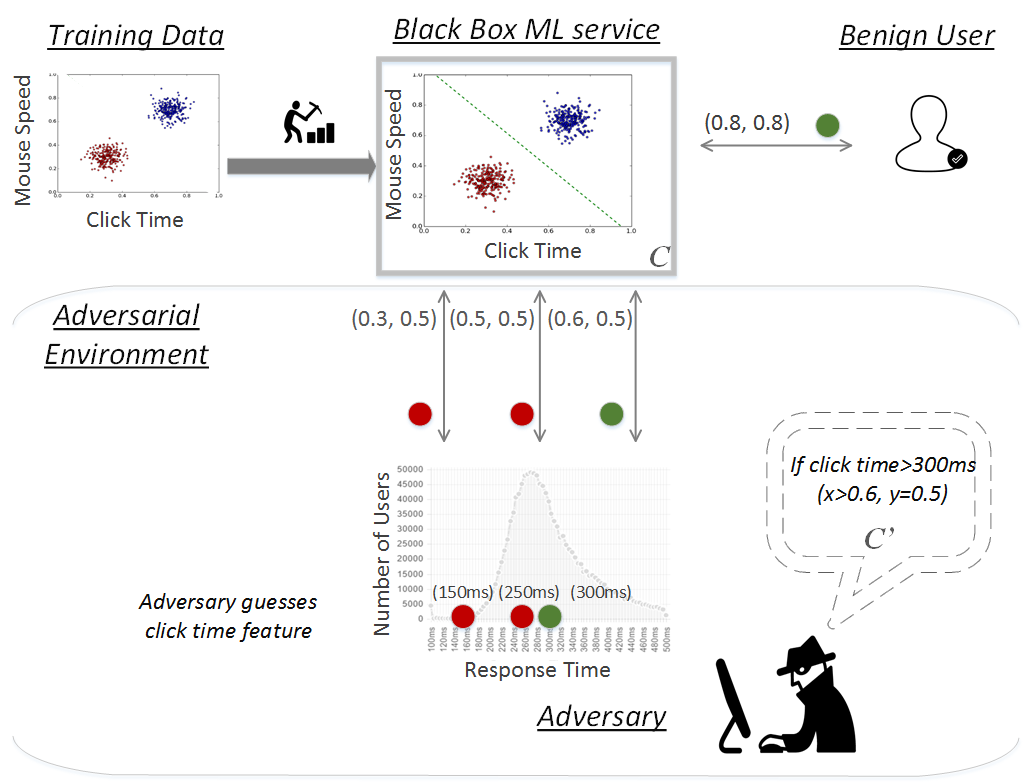}}

\caption{Classifiers in adversarial environment, a) shows the general adversarial nature of the problem and b) shows an example considering a behavioral CAPTCHA system. }
\label{fig:adversarialmodel}
\end{figure}

In an adversarial environment, the accuracy of classification has little significance, if an attacker can easily evade detection by intelligently perturbing the input samples \citep{kantchelian2013approaches}. Any deployed classifier is susceptible to probing based attacks, where an adversary uses the same channel as the client users, to gain information about the system, and then subsequently uses that information to evade detection \citep{abramson2015toward, biggio2014pattern}. This is seen in Fig.~\ref{fig:adversarialmodel} a), where the defender starts by learning from the training data and then deploys the classifier $C$, to provide services to client users. Once deployed, the model $C$ is vulnerable to adversaries, who try to learn the behavior of the defender's classifier by submitting probes as input samples, masquerading as client users. In doing so, the defender's classifier is seen only as a black box, capable of providing tacit \textit{Accept/Reject} feedback on the submitted samples. An adversary, backed by the knowledge and understanding of machine learning, can use this feedback to reverse engineer the model $C$ (as $C'$). It can then avoid detection on future attack samples, by accordingly perturbing the input samples. It was shown recently that, deep neural networks are vulnerable to adversarial perturbations \citep{papernot2016limitations}. A similar phenomenon was shown to affect a wide variety of classifiers in \citep{papernot2016transferability}, where it was demonstrated that adversarial samples are transferable across different classifier families. Cloud based machine learning services (such as Amazon AWS Machine Learning\footnote{\url{https://aws.amazon.com/machine-learning/}} and Google Cloud Platform\footnote{\url{cloud.google.com/machine-learning}}), which provide APIs for accessing predictive analytics as a service, are also vulnerable to similar black box attacks \citep{tramer2016stealing}. 

An example of the aforementioned adversarial environment is illustrated in Fig.~\ref{fig:adversarialmodel}b), where a behavioral mouse dynamics based CAPTCHA (Completely Automated Public Turing test to tell Computers and Humans Apart) system is considered. Popular examples of these systems are Google's reCAPTCHA\footnote{\url{www.google.com/recaptcha}\label{recaptch}} and the classifier based system developed in \citep{d2014avatar}. These systems use mouse movement data to distinguish humans from bots and provide a convenient way to do so, relying on a simple point and click feedback, instead of requiring the user to infer garbled text snippets \citep{d2014avatar}.  The illustrative 2D model of Fig.~\ref{fig:adversarialmodel} b), shows a linear classifier trained on the two features of - Mouse movement speed and Click time. An adversary, aiming to evade detection by this classifier, starts by guessing the click time as a key feature (intuitive in this setting), and then proceeds to makes probes to the black box model $C$, to learn its behavior. Probes are made by going through the spectrum of average reaction times for humans\footnote{\label{benchmark}\url{www.humanbenchmark.com/tests/reactiontime}}, guided by the \textit{Accept}(green)/\textit{Reject}(red) feedback from the CAPTCHA server. The information learned by reconnaissance on the black box system, can then be used to modify the attack payload so as to subsequently evade detection. While, this example was simplistic, its purpose is to illustrate the adversarial environment in which classifiers operate. Practical deployed classifiers tend to be more complex, non linear and multidimensional. However, the same reasoning and approach can be used to evade complex systems. An example of this is the good words and bad words attacks on spam detection systems \citep{lowd2005good}. By launching two spam emails each differing in only one word – \textit{'Sale'}, it can be ascertained that this word is important to the classification, if the email containing that word is flagged as spam. Knowing this information, the adversary can modify the word to be \textit{'Sa1e'}, which looks visually the same but avoids detection. These evasion attacks are non-intrusive in nature and difficult to eliminate by traditional encryption/security techniques, because they use the same access channels as regular input samples and they see the same black box view of the system.  From the classification perspective, these attacks occur at test time and are aimed at increasing the false negative rate of the model, i.e. increase the number of \textit{Malicious} samples classified as \textit{Legitimate} by $C$ \citep{barreno2006can, biggio2013evasion}.

Data driven attacks on deployed classification systems, presents a symmetric flip side to the task of learning from data. Instead of learning from labeled data to generate a model, the task of an attacker is to learn about the model, to generate evasive data samples \citep{abramson2015toward}. With this motivation, we propose the Seed-Explore-Exploit(SEE) framework in this paper, to analyze the attack generation process as a learning problem, from a purely data driven perspective and without incorporating any domain specific knowledge. Research work on detecting concept drift in data streams\citep{tsethi, sethi2016grid}, motivated the need for a formal analysis of the vulnerabilities of machine learning,  with an initial evaluation proposed in our work in \citep{tsethi2016}. In this paper, we extend the earlier version with: i) incorporating and evaluating effects of \textit{Diversity} of attacks on the defender's strategy, ii) introducing adversarial metrics of attack quality and the effects of varying the parameters of attack algorithms, iii) extensive detailed experimentation of the framework using a variety of defender models and the Google Cloud Prediction Service, and iv) experimentation simulating effects of diversity on blacklisting based countermeasures. The main contributions of this paper are:

\begin{itemize}

\item A domain independent data driven framework is presented, to simulate attacks using an Exploration-Exploitation strategy. This generic framework and the algorithms presented, can be used to analyze simple probing attacks to more sophisticated reverse engineering attacks. 
\item Formal adversarial model and adversary's metrics are proposed, as a background for sound scientific development and testing for secure learning frameworks.
\item Empirical analysis on 10 real world datasets, demonstrates that feature space information is sufficient to launch attacks against classifiers, irrespective of the type of classifier and the application domain. Additional experimentation on Google's Cloud Prediction API, demonstrates vulnerability of remote black box prediction services.
\item The analysis of diversity and its effect on blacklisting based countermeasures, demonstrates that such security measures (as proposed in \citep{kantchelian2013approaches}) are ineffective when faced with reverse engineering based attacks of high diversity. 
\end{itemize}

The rest of the paper is organized as follows: Section~\ref{sec:relatedwork}, presents background and related work on the security of machine learning.  Section~\ref{sec:pm}, presents the formal model of an adversary and the proposed Seed-Explore-Exploit(SEE) framework, for attack generation. Based on the framework, the Anchor Points attack  and the Reverse Engineering attack algorithms are presented in Section~\ref{sec:ap} and \ref{sec:re}, respectively. Experimental evaluation and detailed analysis is presented in Section~\ref{sec:expt}. Additional discussion about diversity of attacks and its importance is presented in Section~\ref{sec:diversty}.  Conclusion and avenues for further research are presented in Section~\ref{sec:conclusion}.

\section{Related Work on the Security of Machine Learning}
\label{sec:relatedwork}

One of the earliest works on machine learning security was presented in \citep{barreno2006can}, where a taxonomy of attacks was defined, based on the principles of information security. The taxonomy categorized attacks on machine learning systems along the three axis of: Specificity, Influence and the type of Security Violation, as shown in Table~\ref{tbl:categorization}. Based on the influence and the portion of the data mining process that these attacks affect, they are classified as being either Causative or Exploratory \citep{biggio2014security}. Causative attacks affect the training data while Exploratory attacks affect the model at test time. Specificity of the attacks, refers to whether they affect a set of targeted samples, in order to avoid detection by perturbing them, or if the attacks are aimed towards indiscriminately affecting the system, with no specific pre-selected instances. Based on the type of security violation, the attacks can aim to violate integrity, by gaining unsanctioned access to the system, or can be used to launch a denial of service availability attack.

\begin{table*}[t]
\centering
\caption{Categorization of attacks against machine learning systems}
\label{tbl:categorization}
\begin{tabular}{|l|l|}
\hline
\multirow{2}{*}{\textbf{Influence}} & \begin{tabular}[c]{@{}l@{}}\textit{Causative-} Attacks influence training data,\\  to mislead learning\end{tabular} \\ \cline{2-2} 
 & \begin{tabular}[c]{@{}l@{}}\textit{Exploratory-} Attacks affect test time data, \\                      to evade detection\end{tabular} \\ \hline
\multirow{2}{*}{\textbf{Specificity}} & \textit{Targeted-} Attack affect onyl particular instances \\ \cline{2-2} 
 & \textit{Indiscriminate-} Attacks irrespective of instances \\ \hline
\multirow{2}{*}{\textbf{\begin{tabular}[c]{@{}l@{}}Security \\ violation\end{tabular}}} & \textit{Integrity-} Results in increased false negatives \\ \cline{2-2} 
 & \begin{tabular}[c]{@{}l@{}}\textit{Availability-} Denial of service attacks, \\                      due to increased errors\end{tabular} \\ \hline
\end{tabular}
\end{table*}

Causative attacks aim to mislead training, by poisoning the training set, so that future attacks are easily evaded \citep{li2014causative,biggio2014pattern}. Causative attacks, although severe in effect, can be prevented by careful curation of the training data \citep{li2014improved} and by keeping the training data secure- using database security measures, authentication and encryption. Exploratory attacks are more commonplace, less moderated and can be launched remotely without raising suspicion. These attacks affect the test time data, and are aimed at reducing the system's predictive performance \citep{biggio2014security}. This is done by crafting the attack samples, to evade detection by the defender's model \citep{biggio2013evasion,  lowd2005adversarial}. Once a model is trained and deployed in a cybersecurity application, it is vulnerable to exploratory attacks. These attacks are non intrusive and are aimed at gaining information about the system, which is then exploited to craft evasive samples, to circumvent the system's security. Since, these attacks use the same channel as the client users, they are harder to detect and prevent. 

 Targeted-Exploratory attacks aim to modify a specific set of malicious input samples, minimally, to disguise them as legitimate. Indiscriminate attacks are more general in their goals, as they aim to produce any sample which will result in the defender's model to have a misclassification. Most work on exploratory attacks are concentrated on the targeted case, considering it as a constrained form of indiscriminate attacks, with the goal of starting with a malicious sample and making minimal modifications to it, to avoid detection \citep{biggio2014pattern,biggio2013evasion,xuautomatically,pastrana2014attacks,lowd2005good}. This idea was formalized in \citep{lowd2005adversarial}, where the Minimal Adversarial Cost (MAC) metric, of a genre of classifiers, was introduced to denote the ease with which classifiers of a particular type can be evaded. The hardness of evasion was given in terms of the number of probes needed to obtain a low cost evasive sample. A classifier was considered easy to evade if making a few optimal modifications to a set of samples resulted in a high accuracy of evasion. Work in \citep{nelson2010near} shows that linear and convex inducing classifier are all vulnerable to probing based attacks, and \citep{nelson2012query} presents efficient probing strategies to carry out these attacks. 

\begin{figure}[t]
  \centering
  \includegraphics[width=0.85\linewidth]{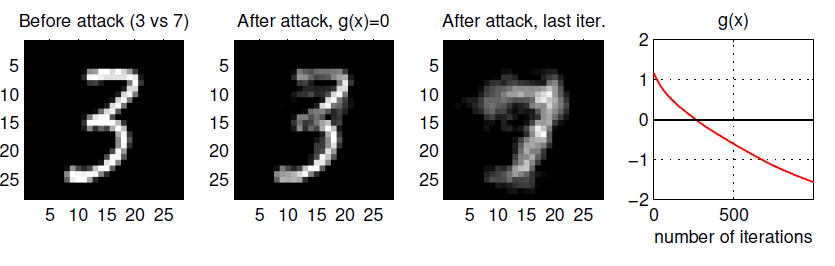}
   \caption{Gradient descent evasion attack over 500 iterations. Left- Initial image of digit 3, Center- Image which first gets classified as 7, Right- Image after 500 iteration.\citep{biggio2013evasion} }
  \label{fig:evasion}
\end{figure}

Particular strategies developed for performing exploratory attacks vary based on the amount of information available to the adversary, with a broad classification presented in \citep{alabdulmohsin2014adding} as: a) Evasion attacks and b) Reverse Engineering attacks. Evasion attacks are used when limited information about the system is available, such as a few legitimate samples only. These legitimate samples are exploited by masking techniques such as- mimicking \citep{smutz2016tree} and spoofing \citep{akhtar2011robustness}, which masquerade malicious content within the legitimate samples. The mimicry attack was presented in \citep{smutz2016tree}, where the Mimicus tool\footnote{\url{www.github.com/srndic/mimicus/blob/master/mimicus/attacks/mimicry.py}} was developed, to implement evasion attacks on pdf documents, by hiding malicious code within benign documents. The good words attacks on spam emails uses a similar technique  \citep{lowd2005good}. A spam email is inserted with benign looking words, to evade detection. Similarly, spoofing attacks are common in biometrics\citep{akhtar2011robustness} and for phishing websites\citep{huh2011phishing}, where visual similarity can be achieved with totally different content. A general purpose, domain independent technique for evasion was presented in \citep{xuautomatically}. Here, using genetic programming, variants of a set of malicious samples were generated as per a monotonically increasing fitness function, which denoted success of evasion. This is an attractive technique due to its generic approach, but limited probing budgets and lack of a graded fitness function, are some of its practical limitations. In the presence of a large probing budget, or specific information about the defender's classifier model, the gradient descent evasion attack of \citep{biggio2013evasion}, can be used. This attacks relies on knowing the exact classifier function used by the defender, or the ability to reverse engineer it using a sufficient number of probes. Once information about the classifier is known, the attack uses a gradient descent strategy to find an optimal low cost evasion sample for the classifier. Search strategies were developed for a wide range of classifiers with differentiable decision functions, including neural networks, non-linear Support vector machines, one class classifiers and for classifiers operating in discrete feature spaces \citep{biggio2013evasion}.  An illustration of the gradient descent attacks for masquerading a sample is shown in Fig.~\ref{fig:evasion}, where the image 3 is modified to be classified as 7 over 500 iterations of gradient descent.

Reverse engineering the defender's model provides avenues for sophisticated exploratory attacks, as it exposes features important to the classifier, to be used for mimicry attacks or large scale indiscriminate attacks. Perfect reverse engineering is not needed, as an adversary is interested only in identifying the portion of the data space which is classified as \textit{Legitimate}. Reverse engineering was first employed in \citep{lowd2005adversarial}, where a sign witness test was used to see if a particular feature had a positive or negative impact on the decision. Reverse engineering of a decision tree classifier, as a symmetric model for defender and adversary, was presented in \citep{xu2014evasion}. \citep{xuautomatically} used genetic programming as a general purpose reverse engineering tool, under the assumption of known training data distribution and feature construction algorithm. The genetic programming output, because of its intuitive tree structure, was then used to guide evasion attacks on intrusion detection systems. The idea of reverse engineering was linked to that of active learning via query generation in \citep{alabdulmohsin2014adding}, where the robustness of SVM to reverse engineering is tested using active learning techniques of random sampling, uncertainty sampling and selective sampling. 

In the above mentioned works of targeted-exploratory attacks, it is assumed that if an evasion is expensive (far from the original malicious sample), the adversary will give up. The above techniques are not designed for a determined adversary, who is willing to launch indiscriminate attacks. An adversary who wants to launch an indiscriminate attack will not bother with the near optimal evasion problem \citep{nelson2010near}. These type of attacks have been largely ignored, with the only mention we found was in \citep{zhou2012adversarial}, where it is termed - the free range attack, as an adversary is free to move about in the data space. In such attacks, the adversary will first analyze the vulnerabilities of the model, looking for prediction blind spots, before attempting an attack. Analyzing performance of models under such attack scenarios is essential to understanding its vulnerabilities in a more general real world situation, where all types of attacks are possible. Also, while most recent methodologies develop attacks as an experimental tool to test their safety mechanisms, there is very few works \citep{biggio2013evasion, xu2014evasion, wagner2002mimicry, pastrana2014attacks}, which have attempted to study the attack generation process itself. Our proposed work analyzes \textit{Indiscriminate-Exploratory-Integrity violating} attacks, under a data driven framework, with diverse adversarial goals and while considering only a black box model for the defender's classifier. We analyze the attacks from an adversary's point of view, considering the adversarial samples generation process, so as to understand the vulnerabilities of classifiers and to motivate the development of secure machine learning architectures.

\section{Proposed Methodology for Simulating Data Driven Attacks on Classifiers}
\label{sec:pm}

Data driven exploratory attacks on classifiers, affect the test time data seen by a deployed classifier. An adversary intending to evade classification, will begin by learning information about the system, over time, and then will launch an attack campaign to meet its goals. The adversary can only interact with the system as a black box model, receiving tacit \textit{Accept/Reject} feedback for the submitted samples. However, an adversary can only make a limited number of probes, before it gets detected or runs out of resources. Additionally, we assume a minimal shared knowledge environment between the adversary and the defender \citep{rndic2014practical}. Only the feature space information is shared between the two parties, as both operate on the same data space. All other information - such as the defender's classifier type, the model parameters and the training data, is kept hidden by the defender. Both the adversary and the defender are assumed to be capable machine learning experts, who are equipped with the tools and understanding of using a data driven approach to best suit their goals. Based on this intuitive understanding, the formal model of an adversary based on it's knowledge, goals and resources \citep{biggio2014pattern}, is presented below:

\begin{itemize}
\item \textbf{Knowledge-} The adversary is aware of the number, type and range of features, used by the classification model. This could be approximated from publications, publicly available case studies in related applications, or by educated guessing \citep{rndic2014practical}. For example, in case of spam classification, the feature domain could be the dictionary of English words, which is publicly available and well known. This represents a symmetric learning problem with both parties operating on the same feature space. No other information about the defender's model is known by the adversary.
\item \textbf{Goals-} The adversary intends to cause false negatives for the defender's classifier, on the submitted attack samples. Additionally, the adversary also wants the attacks to be robust, such that it can avoid being detected and stopped by simple blacklisting techniques \citep{bilge2012before}. From a data driven perspective, the attacker aims to avoid detection by generating an attack set with high diversity and variability. While, repeating a single confirmed attack point, over and over, leads to ensured false negatives, such attacks are easily stopped by blacklisting that single point. We consider serious attackers only, who aim to force the retraining of the defender's classification system. 
\item \textbf{Resources-} The adversary has access to the system only as a client user. It can submit probes and receive binary feedback on it, upto a limited probing budget, without being detected. The adversary does not have control over the training data or the model trained by the defender. 
\end{itemize}

The presented model of the adversary represents a general setting, where an attacker can take the form of an end user and then attack the system over time. Modern day web applications, which aim to reach as many users as possible, all operate under this environment and are susceptible to data driven attacks. Based on the adversary's model, the attack can be formalized here. A classifier $C$, trained on a set of training data $D_{Train}$, is responsible for classifying incoming samples into \textit{Legitimate} or \textit{ Malicious} classes. An adversary aims to generate an attack campaign of samples $D'_{Attack}$, such that $C(D'_{Attack})$ has a high false negative rate.  The adversary has at its disposal, a budget $B_{Explore}$ of probing data $D'_{Explore}$, which it can use to learn $C$ and understand it as $C'(D'_{Explore})$. The number of attack samples ($N_{Attack}$) should be much larger than $B_{Explore}$, to justify expenditure on the adversary's part. This specified notation will be used through the rest of the paper.

 The Seed-Explore-Exploit (SEE) framework is presented in Section~\ref{sec:see}, which provides an overview of the attack paradigm. Two specific attack strategies developed under the SEE framework, the Anchor Points attacks (AP) and the Reverse Engineering attacks (RE), are presented in Section~\ref{sec:ap} and Section~\ref{sec:re}, respectively. 

\subsection{The Seed-Explore-Exploit (SEE) Framework}
\label{sec:see}

The SEE framework employs a data driven approach for generating adversarial samples. The idea of Exploration-Exploitation is common in search based optimization techniques, where the goal is to learn the data space and then emphasize only on the promising directions \citep{chen2009optimal}. An adversary can also utilize a similar strategy, to best utilize the exploration budget ($B_{Explore}$), such that the resulting attack samples ($D'_{Attack}$) have high accuracy and high diversity. The specific steps of the framework are explained below:

\begin{itemize}

\item \textbf{Seed-} An attack starts with a seed phase, where it acquires a legitimate sample (and a malicious sample), to form the seed set $D'_{Seed}$. This seed sample can be acquired by random sampling in the feature space, by guessing a few feature values, or from an external data source of a comparable application \citep{papernot2016transferability}. For the case of a spam classification task, picking an email from one's own personal inbox would be a functional legitimate seed sample.  
\item \textbf{Explore-} Exploration is a reconnaissance task, starting with $D'_{Seed}$, where the goal is to obtain maximum diverse information, to understand the coverage and extent of the space of legitimately classified samples. In this phase, the adversary submits probes and receives feedback from the defender's black box. The defender can be probed upto a budget $B_{Explore}$, without being thwarted or detected. To avoid detection, it is natural that the adversary needs to spread out the attacks over time and data space, in which case the $B_{Explore}$ is the time/resources available to the adversary. The exploration phase results in a set of labeled samples $D'_{Explore}$, and the goal of the adversary is to best choose this set based on it's strategy. 
\item \textbf{Exploit-} The information gathered in the exploration phase is used here to generate a set of attack samples $D'_{Attack}$. The efficacy of the attack is based on the accuracy and the diversity of these samples. 
\end{itemize}

The SEE framework provides a generic way of defining attacks on classifiers. Specific instantiations of the three phases can be developed, to suit one needs and simulation goals.

\subsection{The Anchor Points Attack (AP)}
\label{sec:ap}
The Anchor Points attack is suited for adversaries with a limited probing budget $B_{Explore}$, who have a goal of generating evasive samples for immediate benefits. An example of this would be - zero day exploits, where an adversary wants to exploit a new found vulnerability, before it is fixed \citep{bilge2012before}. These attacks start by obtaining a set of samples classified as Legitimate by $C$, called the Anchor Points, which serve as ground truth for generating further attack samples. From a data driven perspective, this attack  strategy is defined under the SEE framework as given below.

\begin{figure*}[t]
\centering
\includegraphics[width=0.85\textwidth]{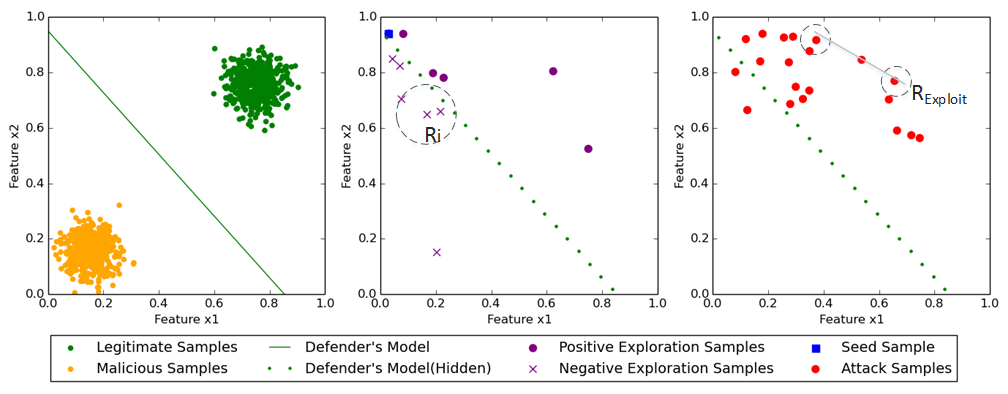}
\caption{Illustration of AP attacks on 2D synthetic data.\textit{(Left - Right)}: The defender's model from it's training data. The Exploration phase depicting the seed(blue) and the anchor points samples(purple). The Exploitation attack phase samples (red) generated based on the anchor points.}
\label{fig:ap}
\end{figure*}

\begin{itemize}
\item \textbf{Seed-} The attack begins with a single legitimate sample as the Seed ($D'_{Seed}$).
\item \textbf{Explore-} After the initial seed has been obtained (provided or randomly sampled), the exploration phase proceeds to generate the set of Anchor Points, which will enable the understanding of the space of samples classified as \textit{Legitimate}. The exploration phase is described in Algorithm~\ref{algo:ap_explore}, and is a radius based incremental neighborhood search technique, around the seed samples, guided by the feedback from the black box model $C$. Diversity of search is maintained by dynamically adjusting the search radius ($R_i$), based on the amount of ground truth obtained so far (Line 5). This ensures that radius of exploration increases in cases where the number of legitimate samples obtained is high, and vice versa, thereby balancing diversity of samples with their accuracy. Samples are explored by perturbing an already explored legitimate sample (Seed sample in case of first iteration), within the exploration radius (Line 7). The final exploration dataset of Anchor Points - $D'_{Explore}$, is comprised of all explored samples $x_i$, for which $C(x_i)$ indicated the \textit{Legitimate} class label. The exploration phase is illustrated on a synthetic 2D dataset in Fig.~\ref{fig:ap}, where the neighborhood radius $R_i$ indicates the exploration neighborhood of a sample. 
\item \textbf{Exploit-} The anchor points obtained as $D'_{Explore}$, forms the basis for launching the dedicated attack campaign on the classifier $C$. The exploitation phase (Algorithm~\ref{algo:ap_exploit}) combines two techniques to ensure high accuracy and diversity of attack samples:  \textit{a) Simple perturbation-} The anchor point samples are perturbed, similar to the exploration phase, using a radius of exploitation- $R_{Exploit}$ (Line 4) and,  \textit{b) Convex combination-} The perturbed samples are combined using convex combination of samples, two at a time (Line 7).  This is inspired by the Synthetic Minority Oversampling Technique (SMOTE), which is a popular oversampling technique for imbalanced datasets \citep{chawla2002smote}. The attack set $D'_{Attack}$, shown in red in Fig.~\ref{fig:ap}, is the final attack on the classifier $C$.
\end{itemize}

\begin{algorithm}[t]
\SetKwInOut{Input}{Input}
\SetKwInOut{Output}{Output}
 \Input{Seed Data $D'_{Seed}$, Defender black box $C$. \textit{Parameters}: Exploration budget $B_{Explore}$,  Exploration neighborhood- [$R_{min}$, $R_{max}$]  }
 \Output{Exploration data set $D'_{Explore}$}
   \setcounter{AlgoLine}{0}

 $D'_{Explore} \leftarrow D'_{Seed}$
 
 count\_ legitimate=0

 \For{ i = 1 .. $B_{Explore}$}{
 
 	$x_i$ $\leftarrow$ Select random sample from $D'_{Explore}$
 	
 	$R_{i} = (R_{max}-R_{min}) * (count\_legitimate / i) + R_{min}$\\ \Comment{\textbf{Dynamic neighborhood search}}
 	
 	$\hat{x_i}\leftarrow$ \textit{Perturb}($x_i$ , $R_{i}$) \Comment{perturbed sample}\\

	\If{C.predict($\hat{x_i}$) is \textit{Legitimate}}
	{
		$D'_{Explore}\cup$ $\hat{x_i}$
		
		count\_legitimate ++
	}
	}   
\textbf{Procedure} Perturb(sample, $R_{Neigh}$)

$\quad$return sample+=random(mean=0, std=$R_{Neigh}$)
\caption {AP- Exploration Phase}
\label{algo:ap_explore}
\end{algorithm}

\begin{algorithm}[t]
\SetKwInOut{Input}{Input}
\SetKwInOut{Output}{Output}
 \Input{Exploration data set $D'_{Explore}$, Number of attacks $N_{Attack}$, Radius of Exploitation $R_{Exploit}$  }
 \Output{Attacks set $D'_{Attack}$}

 $D'_{Attack}\leftarrow$[]

 \For{ i = 1 .. $N_{Attack}$}{
 
 		$x_A,x_B \leftarrow$ Select random samples from $D'_{Explore}$
 		 				
 		$\hat{x_A},\hat{x_B}\leftarrow Perturb(x_A,R_{Exploit})$, $Perturb(x_B,R_{Exploit})$ \\ \Comment{\textbf{Random perturbation}}
 		
		$\lambda=random(0,1)$ \Comment{number in [0,1]}

		$attack\_sample_i \leftarrow \hat{x_A}* \lambda + (1-\lambda)*\hat{x_B}$
		\\	 \Comment{\textbf{Convex combination}}
					
		$D'_{Attack}\cup attack\_sample_i$
	}

\textbf{Procedure} Perturb(sample, $R_{Exploit}$)

$\quad$return sample+=random(mean=0, std=$R_{Exploit}$)
\caption {AP- Exploitation Phase}
\label{algo:ap_exploit}
\end{algorithm}

The performance of the AP attack is largely dependent on the probes collected in the initial seed and exploration phase. As such, maintaining diversity is key, as larger coverage ensures more flexibility in attack generation. By the nature of these attacks, they can be thwarted by blacklists capable of approximate matching \citep{prakash2010phishnet}. Nevertheless, they are suited for adhoc swift blitzkriegs, before the defender has time to respond. 

\subsection{The Reverse Engineering Attack (RE)}
\label{sec:re}

In case of sophisticated attackers, with a large $B_{Explore}$, direct reverse engineering of the classification boundary is more advantageous. It provides a better understanding of the classification landscape, which can then be used to launch large scale evasion or availability attacks \citep{kantchelian2013approaches}. Reverse engineering could also be an end goal in itself, as it provides information about feature importance to the classification task \citep{lowd2005adversarial}. A reverse engineering attack, if done effectively, can avoid detection and make retraining harder on the part of the defender. However, unlike the AP attacks, these attacks are affected by the type of model used by the black box $C$, the dimensionality of the data and the number of probes available. Nevertheless, the goal of an adversary is not to exactly fit the decision surface, but to infer it sufficiently, so as to be able to generate attacks of high accuracy and diversity. As such, a linear approximation to the defender's model and a partial reverse engineering attempt should be sufficient for the purposes of launching a reduced accuracy attack. This reduction in accuracy can be compensated for by launching a massive attack campaign, exploiting the information provided by the reverse engineered model $C'$.

Effective reverse engineering relies on the availability of informative samples. As such, it is necessary to use the probing budget $B_{Explore}$ effectively. Random sampling can lead to wasted probes, with no additional information added, making it ineffective for the purposes of this attack. The query synthesis technique of \citep{wang2015active}, generates samples close to the classification boundary and spreads the samples along the boundary, to provide a better learning opportunity. The approach of \citep{wang2015active} was developed for the purpose of active labeling of unlabeled data. We modify the approach to be used for reverse engineering as part of the SEE framework, where  the attacker learns a surrogate classifier $C'$, based on probing the defender's black box $C$. The SEE implementation of the RE attack is given below:

\begin{algorithm}[t]
\SetKwInOut{Input}{Input}
\SetKwInOut{Output}{Output}
 \Input{Seed Data $D'_{Seed}$, Defender black box model $C$. \textit{Parameters}: Exploration budget $B_{Explore}$, Magnitude of dispersion $\lambda_{max}$}
 \Output{Exploration data Set $D'_{Explore}$, Surrogate classifier $C'$}

 $D'_{Explore\_L}$= Legitimate samples of $D'_{Seed}$
 
 $D'_{Explore\_M}$= Malicious samples of $D'_{Seed}$

 \For{ i = 1 .. $B_{Explore}$}{
 		$x_L\leftarrow$ Select random samples from  $D'_{Explore\_L}$

 	 	$x_L\leftarrow$ Select random samples from  $D'_{Explore\_M}$
 	 	
 	 	$x_0$=	$x_L-x_M$
 	 	
 	 	Generate random vector $x_R$
 	 	
		$x_R=x_R-\frac{<x_R,x_0>}{<x_0, x_0>}*x_0$  \\
		\Comment{\textbf{Gram-Schmidt process - $x_R$ orthogonal to $x_0$}}
		
		$\lambda_i=random(0,\lambda_{max})$		
		
		$x_R$=$\frac{\lambda_i}{norm(x_R)}$*$x_R$ \\
		\Comment{set magnitude of orthogonal midperpendicular}
		
		$x_S$=$x_R+(x_L+x_M)/2$ \Comment{Set $x_R$ to midpoint}
		
		\If{C.predict($x_S$) is Legitimate}{
		$D'_{Explore\_L} \quad \cup \quad x_S$
		}
		\Else{
				$D'_{Explore\_M} \quad \cup \quad x_S$
		}
	}
	$D'_{Explore}$ =	$D'_{Explore\_L} \quad \cup\quad D'_{Explore\_M}$
	
	Train $C'$ using $D'_{Explore}$ \\
	\Comment{Training can be based on linear classifier of choice}   

\caption {RE Exploration - Using Gram-Schmidt process.}
\label{algo:re_explore}
\end{algorithm}

\begin{figure*}[t]
\centering
\includegraphics[width=0.85\textwidth]{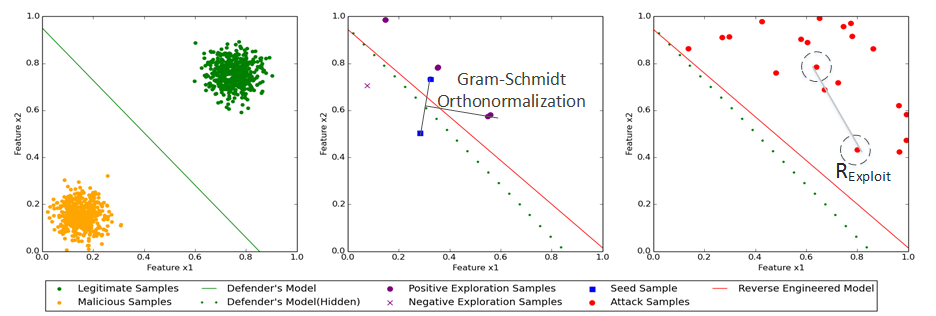}
\caption{Illustration of RE attacks on 2D synthetic data.\textit{(Left - Right)}: The defender's model based on training data. The Exploration phase depicting reverse engineering(red) using the Gram-Schmidt orthonormalization process. The Exploitation attack phase samples generated after validation from the surrogate classifier (red samples).}
\label{fig:re}
\end{figure*}

\begin{itemize}

\item \textbf{Seed-}  The seed set consists of one legitimate and one malicious class sample. 
\item \textbf{Explore-} The exploration phase (Algorithm~\ref{algo:re_explore}) uses the Gram-Schmidt process \citep{wang2015active} to generate orthonormal samples, near the midpoint of any two randomly selected seed points of opposite classes (Line 8). This has the effect of generating points close to the separating decision boundary of the two classes, and also of spreading the samples along this boundary's surface, as depicted in the exploration phase of Fig.~\ref{fig:re}. The magnitude of the orthonormal vector is set based on $\lambda_i$, which is selected as a random value in [0,$\lambda_{max}$], to impart diversity to the obtained set of samples (Line 10-11). At the end of the exploration phase, the resulting set of labeled samples ($D'_{Explore}$), is used to train a linear classifier of choice, to form the surrogate reverse engineered model $C'$ (Line 19). Fig.~\ref{fig:re} shows the reverse engineered model (red), as learned from the original black box classifier $C$ (green). 

\item \textbf{Exploit-} The surrogate model $C'$, can be used to generate attacks with high accuracy and diversity. Ideally, a set of random points can be generated and verified against the reverse engineered model $C'$, before adding them to the attack set $D'_{Attack}$. However, a practical and efficient way would be to use the exploration set samples $D'_{Explore}$ of Algorithm~\ref{algo:re_explore}, as a seed set to generate a set of anchor points as in Algorithm~\ref{algo:ap_explore}, with the exception that we probe $C'$ instead of the original model $C$. Since $C'$ is a locally trained model, probing it does not impact $B_{Explore}$. Thus allowing an adversary to make a large number of probes, at theoretically zero cost. The anchor points obtained can then be used to generate the attack samples using Algorithm~\ref{algo:ap_exploit}. A larger attack radius $R_{Exploit}$ can be used with this attack strategy, as additional validation is available via the model $C'$. 
\end{itemize}

The RE attack strategy is suited for a patient adversary, who spends time/effort to probe the system and learn it, so as to have an effective attack with high diversity. Such attacks, are often hard to detect and stop by simple blacklisting techniques. However, the success of this attack relies on the goodness of the reverse engineered model, and could be affected by the nature of learning employed by the black box.

\section{Experimental Evaluation}
\label{sec:expt}

This section presents experimental evaluation of the AP and the RE approaches, on classifiers trained with 7 real world datasets. Additionally, evaluation on 3 datasets from the cybersecurity domain is presented, to demonstrate the vulnerabilities of machine learning systems in adversarial milieus, to exploratory attacks. The experiments are presented from an adversary's point of view, who wishes to have effective attacks with high accuracy and diversity. Section~\ref{sec:exptmethods} presents the metrics and the experimental protocol used, to encourage reproduce-ability of results. Experimental results and discussions is presented in Section~\ref{sec:results}

\subsection{Experimental Methods and Setup}
\label{sec:exptmethods}

\subsubsection{Adversary Metrics for Attack Quality}
\label{sec:metrics}

An adversary aiming to create maximum impact, needs to make the set of attack samples $D'_{Attack}$ - \textit{Accurate} and \textit{Diverse}. Accuracy ensures that the attack samples will cause an increase in the false negative rate of the defender's model $C$. While, diversity ensures that the attack set has enough variability, so that they can go unnoticed for a long time. These intuitive ideas are quantified using 4 proposed quality metrics, to measure adversary effectiveness. The Effective Attack Rate (EAR) measures the accuracy of attacks, and the 3 metrics: Deviation of attacks ($\sigma_{EA}$), K-Nearest Neighbor Distance (KNN-dist) and the Minimum Spanning Tree distance (MST-dist), collectively represent the diversity of attacks. The metrics are based on the following definition of effective attacks (EA):

\begin{equation}
EA\quad =\quad \left\{ x:\quad C(x) = Legitimate ~ \wedge ~ x\in D'_{Attacks} \right\} 
\label{eqn:EA}
\end{equation}

Based on Eqn.~\ref{eqn:EA}, an attack sample is effective if it is classified as \textit{Legitimate} by the black box $C$. The adversarial quality metrics over the set EA are defined below:

\begin{enumerate}[label=\alph*]
\item) \textit{Effective Attack Rate (EAR)}: This is the accuracy of attacks, measured as the ratio of attack samples which successfully evade the defenders classifier, given by Eqn.~\ref{eqn:EAR}. A value of 1 denotes perfect evasion. 
\begin{equation}
EAR\quad =\quad \frac { \left| EA \right|  }{ \left| D'_{Attacks} \right|  } 
\label{eqn:EAR}
\end{equation}

\item) \textit{Deviation of effective attacks ($\sigma_{EA}$)}: This is a measure of diversity, which computes the spread of data around its mean, given by Eqn.~\ref{eqn:std_ea}.   
\begin{equation}
{ \sigma  }_{ EA }\quad =\quad \sqrt { \frac { 1 }{ \left| EA-1 \right|  } \sum _{ x_i\epsilon EA }{ { \left( { x_i }-{ \mu  }_{ EA } \right)  }^{ 2 } }  } 
\label{eqn:std_ea}
\end{equation}

where, $\mu_{EA}$ indicates the Euclidean mean of samples in the effective attack set EA. A large value of $\sigma_{EA}$ indicates that the data has high data space coverage.

\item) \textit{K-Nearest Neighbor distance of effective attacks ($KNN-dist_{EA}$)}: This measure of diversity, computes local density information of the data samples (motivated by \citep{he2007nearest}). It is computed by finding the average distance of the K-nearest neighbors of a sample, for all samples and then  averaging this value, as given by Eqn.~\ref{eqn:knn}. 
\begin{equation}
{ KNN-dist }_{ EA }\quad =\quad \frac { \sum _{ x\epsilon EA }^{  }{ \sum _{ i=1 }^{ K }{ \quad dist(x,{ NN }_{ i }(x)) }  }  }{ K.\quad \left| EA \right|  } 
\label{eqn:knn}
\end{equation}

Where, the \textit{dist(.)} function computes Euclidean distance between two vectors, and $NN_i(x)$ gives the $i^{th}$ nearest neighbor of a sample \textit{x}. A higher value of KNN-dist, indicates that data samples are relatively far from each other and that every sample is in a locally sparse region of space, indicating higher spread. A value of K=5 is chosen for experimentation.

\item) \textit{Minimum Spanning Tree distance of effective attacks ($MST-dist_{EA}$)}: This is also a measure of diversity, which is computed by finding the length of the minimum spanning tree over the set of EA samples, as per Eqn.~\ref{eqn:mst} \citep{lacevic2011ectropy}. This is a measure which promotes ectropy or collocation of points, in an attempt to obtain a more global uniform and diverse spread of samples. This is especially useful in recognizing multiple locally dense clusters which are far from each other. 
\begin{equation}
{ MST-dist }_{ EA }\quad =\quad \frac { length(MST(EA)) }{ \left| EA \right| -1 } 
\label{eqn:mst}
\end{equation}

The MST measure computes cluster separation only once, as opposed to pairwise distance metrics which calculate distance between one point and every other point. Thus the MST provides a better sense of global diversity, by allowing sub groups of data to have less diversity. A high value of MST-distance will indicate high diversity.  
\end{enumerate}

\begin{figure*}[t]
\centering
\subfloat[$\sigma$=0.228; KNN-dist=0.071; MST-dist=0.056]{\includegraphics[width=0.24\linewidth]{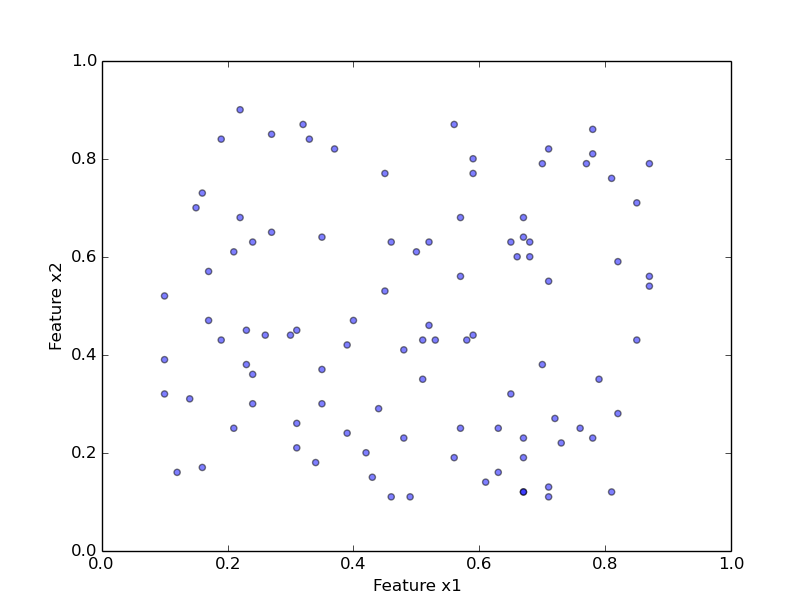}}
\subfloat[$\sigma$=0.058; KNN-dist=0.018; MST-dist=0.015]{\includegraphics[width=0.24\linewidth]{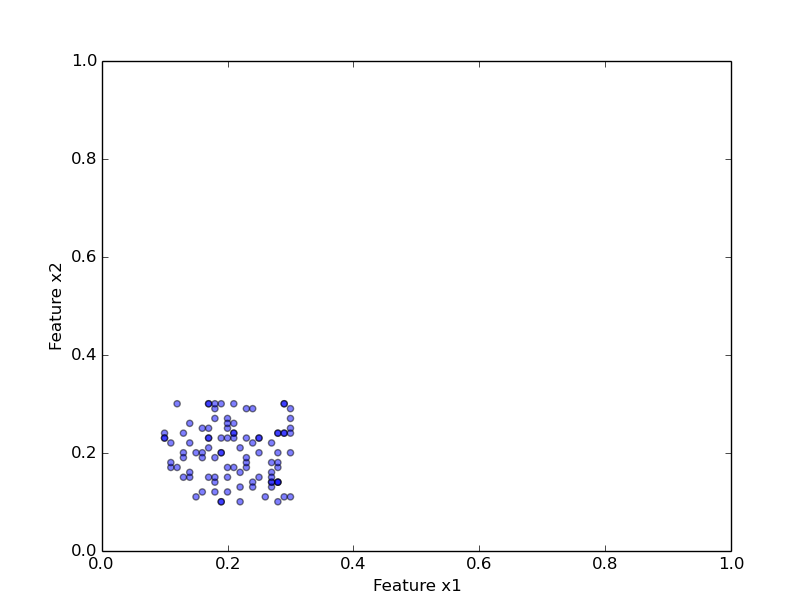}}
\subfloat[$\sigma$=0.218; KNN-dist=0.017; MST-dist=0.037]{\includegraphics[width=0.24\linewidth]{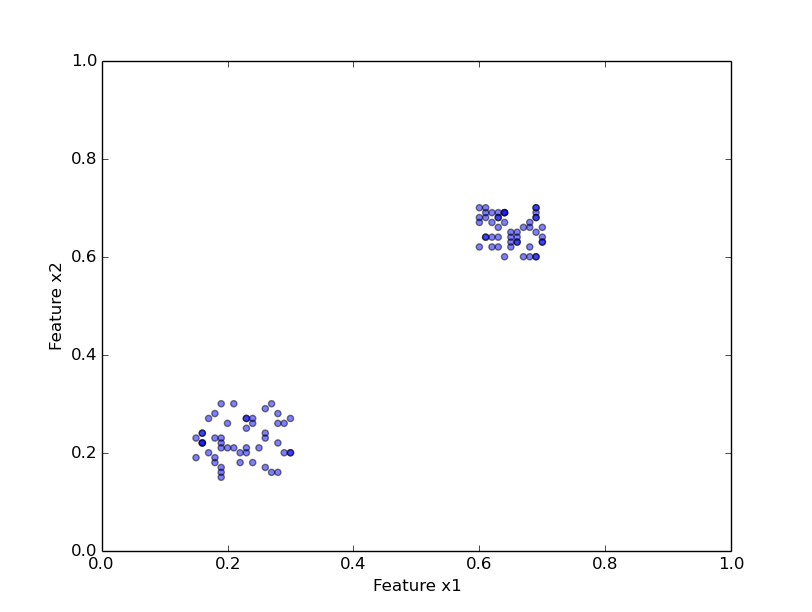}}
\subfloat[$\sigma$=0.226; KNN-dist=0.017; MST-dist=0.063]{\includegraphics[width=0.24\linewidth]{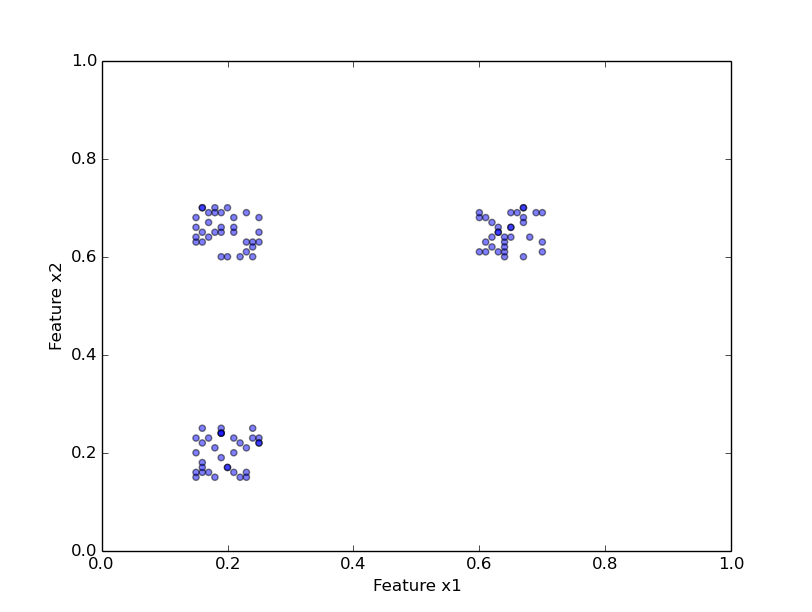}}
\caption{Values of $\sigma$, \textit{KNN-dist} and \textit{MST-dist} for different 2D synthetic data distributions over 100 test points.}
\label{fig:metrics}
\end{figure*}

The three diversity metrics are affected by different data distributions and together they provide a holistic representation of the variability of the attack data. Standard deviation captures the overall spread of the data and is severely affected by outliers. A larger spread of data results in higher deviation, as seen in Fig.~\ref{fig:metrics} a) where the deviation $\sigma$=0.228 is higher than in b), where the deviation is $\sigma$=0.058. Although, deviation is effective in capturing the global spread of data, it fails at capturing the local data characteristics. As seen in Fig.~\ref{fig:metrics} a) and c), which have very close deviation values ($\Delta\sigma$=0.002), but totally different distribution of data. These differences are caught by the $KNN-dist$ metric, which is higher for scattered data (Fig.~\ref{fig:metrics} a), \textit{KNN-dist}=0.071), as compared to closely packed data (Fig.~\ref{fig:metrics} c), \textit{KNN-dist}=0.017). However, the \textit{KNN-dist} metric does not account for disjoint clusters spread out in space, as its a local measure and is myopic in scope. This distinction is caught effectively by the \textit{MST-dist} metric, which shows a significant difference in the diversity values for Fig.~\ref{fig:metrics} c) and d) ($\Delta MST-dist$=0.026), even though the \textit{KNN-dist} metric shows no difference between the two. The MST metric is suitable for attacks such as the Anchor Points attacks, where the attacks are concentrated around a few ground truth points, but the ground truth points themselves are spread out in space. The three metrics together represent the variability of the samples, in high dimensional spaces, where a visual examination of the data is not possible. The subscript $EA$ is  omitted in the representation of the diversity metrics, through the rest of the paper, with the implicit understanding that these metrics are computed over the effective attacks set only.

\subsubsection{Description of Datasets Used}
\label{sec:datasets}

Experimental evaluation is performed on 10 real world datasets, the details of which are presented in Table~\ref{tbl:datasets}. The first 7 datasets were chosen from the UCI machine learning \citep{Lichman:2013} repository and are popularly used for classification tasks, in literature. These datasets do not traditionally embody any security risks, but were chosen to evaluate the vulnerability of classifiers in the different data domains and distributions. The Spambase\footnote{\url{https://archive.ics.uci.edu/ml/datasets/Spambase}}\citep{Lichman:2013}, KDD99\footnote{\url{http://kdd.ics.uci.edu/databases/kddcup99/kddcup99.html}}\citep{Lichman:2013} and the CAPTHCA\ \citep{d2014avatar} datasets are binary classification tasks, which represent 3 different cybersecurity domains that use machine learning as a core technique. The Spambase dataset, contains data about spam emails (such as fraud schemes, ads, etc) and legitimate personal and work emails. The KDD99 dataset is a network intrusion detection dataset, to classify normal connections from different classes of attack connections. The CAPTCHA dataset was developed in \citep{d2014avatar}, for the task of blocking bots from human users, based on their mouse movement patterns, while solving a visual image based behavioral CAPTCHA puzzle. 

\begin{table}[t]
\centering
\caption{Description of datasets used for experimentation of SEE framework}
\label{tbl:datasets}
\begin{tabular}{|l|l|l|}
\hline
Dataset & \#Instances & \#Dimensions \\ \hline
Digits08 & 1500 & 16 \\ \hline
Credit & 1000 & 61 \\ \hline
Cancer & 699 & 10 \\ \hline
Qsar & 1055 & 41 \\ \hline
Sonar & 208 & 60 \\ \hline
Theorem & 3060 & 51 \\ \hline
Diabetes & 768 & 8 \\ \hline
Spambase & 4600 & 57 \\ \hline
KDD99 & 494021 & 41 \\ \hline
CAPTCHA & 1885 & 26 \\ \hline
\end{tabular}
\end{table}

All datasets were pre-processed by first reducing them to a binary class problem. The Digits dataset was reduced to have samples of the digit 0 and 8 only, KDD99 was reduced to represent only two classes - attacks and normal. The dataset was then converted to contain only numerical values by transforming categorical and nominal features to binary variables. The resulting number of features is shown in Table~\ref{tbl:datasets}. The data was then normalized to the range of [0,1]. Instances were shuffled to remove any bias due to inherent concept drift. In all datasets, the class label 1 is taken to be the \textit{Malicious} class and 0 is taken as the \textit{Legitimate} class, as convention.  

\subsubsection{Experimental Protocol and Setup}
\label{sec:setup}
All experiments begin with a seed sample, which is obtained by random sampling in the feature space. The Anchor Points(AP) attack requires only one legitimate seed sample while the Reverse Engineering(RE) attack requires one legitimate and one malicious sample. The seed phase concludes when the minimum required seed samples are obtained. The exploration probing budget $B_{Explore}$ is taken as 1000 samples and the number of attack samples required $N_{Attack}$ is taken as 2000. For the AP attack, the neighborhood radius [$R_{min}, R_{max}$] is set at [0.1,0.5] and the exploitation radius is set at $R_{Exploit}$=0.1. In case of the RE attack, a larger exploitation radius is taken as $R_{Exploit}$=0.5, due to additional validation from the surrogate learned classifier $C'$. Effects of varying this radius values are also presented in the analysis. The magnitude of dispersion $\lambda_{max}$ is taken as 0.25, and it was found that changing this had little impact on the final results. The adversary's reverse engineered model is taken as a linear kernel SVM with a high regularization constant (c=10). This ensures that the model is robust and does not overfit to the explored samples, which are limited and inadequate to generalize over the entire space. All experimentation was performed using Python 2.7\footnote{\url{www.python.org}} and the scikit-learn machine learning library \citep{scikit-learn}. The results presented are averaged over 30 runs for every experiment.

\subsection{Experimental Results and Analysis}
\label{sec:results}

Experimental analysis is presented here, by considering different models for the defender's black box, and measuring its impact on the adversary's effectiveness. Section~\ref{sec:linear} presents the results of a symmetric case, where both the adversary and the defender have similar model types (linear in this case). Results of a non symmetric setting are presented in Section~\ref{sec:nonlinear}, where we consider 4 different model types for the defender, while the adversary, agnostic of these changes, still employs a linear model. Experiments on a truly remote black box model is presented in Section~\ref{sec:gcp}, where we present experiments performed on Google's Cloud Prediction API. Effects of parameters on the adversary's performance is presented in Section~\ref{sec:parameters}.

\subsubsection{Experiments with Linear Defender Model}
\label{sec:linear}

Experiments in this section consider a linear model for the defender's classifier $C$. A linear kernel SVM (regularization parameter,  c=1) is considered. This information is not available to the adversary, who is capable of accessing this model only via probing upto a budget $B_{Explore}$=1000.

The results of the \textit{Seed} and \textit{Exploration} phase are presented in Table~\ref{tbl:real_world_se}. The initial accuracy of the defender, as perceived by cross-validation on its training dataset before deployment, is seen in Column 2 of Table~\ref{tbl:real_world_se}. A high accuracy ($>70\%$) is seen across all the datasets. The seed phase uses random sampling in the feature space to find seed samples. No more than 50 samples, on average, were needed for finding seeds to start the attack process. The number of Anchor Points obtained is seen to be $>50\%$ of $B_{Explore}$, indicating the ability to launch an AP attack on all 10 high dimensional domains. For the RE attack, the reverse engineering accuracy of model $C'$ is computed by evaluating it on the original dataset, as an adhoc metric of $C'$'s understanding of the original data space and the extent of reverse engineering. 

\begin{table*}[t]
\centering
\caption{Results of Seed and Exploration phases, with linear defender model}
\label{tbl:real_world_se}
\begin{tabular}{|l|c|c|c|c|}
\hline
Dataset & \begin{tabular}[c]{@{}c@{}}Defender's\\   Initial Accuracy\end{tabular} & \begin{tabular}[c]{@{}c@{}}Random probes\\ to find seed\end{tabular} & \begin{tabular}[c]{@{}c@{}}Explored Anchor\\ Points/$B_{Explore}$\end{tabular} & \begin{tabular}[c]{@{}c@{}}Accuracy of\\ RE model $C'$\end{tabular} \\ \hline
Digits08 & 98\% & 4.6$\pm$2.63 & 0.63$\pm$0.01 & 92\% \\ \hline
Credit & 79\% & 3.13$\pm$1.89 & 0.71$\pm$0.01 & 71\% \\ \hline
Cancer & 97\% & 42.91$\pm$29.36 & 0.99$\pm$0.01 & 95\% \\ \hline
Qsar & 87\% & 49.5$\pm$28.81 & 0.99$\pm$0.01 & 42\% \\ \hline
Sonar & 88\% & 24.03$\pm$18.92 & 0.98$\pm$0.01 & 61\% \\ \hline
Theorem & 72\% & 4.07$\pm$2.52 & 0.67$\pm$0.02 & 57\% \\ \hline
Diabetes & 78\% & 2.93$\pm$1.23 & 0.50$\pm$0.02 & 71\% \\ \hline
Spambase & 91\% & 20.64$\pm$12.93 & 0.50$\pm$0.02 & 59\% \\ \hline
KDD99 & 99\% & 6.07$\pm$4.23 & 0.91$\pm$0.01 & 55\% \\ \hline
CAPTCHA & 100\% & 7.27$\pm$5.35 & 0.92$\pm$0.01 & 91\% \\ \hline
\end{tabular}
\end{table*}

\begin{table*}[t]
\centering
\caption{Results of accuracy and diversity of AP and RE attacks, with linear defender model}
\label{tbl:real_word_exploit}
\begin{tabular}{|l|c|c|c|c|c|}
\hline
Dataset & Method & EAR & $\sigma$ & KNN-dist & MST-dist \\ \hline
\multirow{2}{*}{Digits08} & AP & 0.96$\pm$0.01 & 0.23$\pm$0.002 & 0.48$\pm$0.01 & 0.41$\pm$0.01 \\ \cline{2-6} 
 & RE & 0.93$\pm$0.06 & 0.273$\pm$0.009 & 0.76$\pm$0.01 & 0.65$\pm$0.04 \\ \hline
\multirow{2}{*}{Credit} & AP & 0.98$\pm$0.01 & 0.218$\pm$0.001 & 1.19$\pm$0.02 & 1.01$\pm$0.02 \\ \cline{2-6} 
 & RE & 0.80$\pm$0.15 & 0.265$\pm$0.001 & 2.22$\pm$0.02 & 1.72$\pm$0.31 \\ \hline
\multirow{2}{*}{Cancer} & AP & 0.99$\pm$0.01 & 0.215$\pm$0.001 & 0.38$\pm$0.01 & 0.33$\pm$0.01 \\ \cline{2-6} 
 & RE & 0.99$\pm$0.01 & 0.263$\pm$0.001 & 0.5$\pm$0.01 & 0.45$\pm$0.01 \\ \hline
\multirow{2}{*}{Qsar} & AP & 1 & 0.216$\pm$0.001 & 1.1$\pm$0.01 & 0.94$\pm$0.01 \\ \cline{2-6} 
 & RE & 0.99+0.01 & 0.264$\pm$0.001 & 1.71$\pm$0.01 & 1.64$\pm$0.01 \\ \hline
\multirow{2}{*}{Sonar} & AP & 0.99$\pm$0.01 & 0.215$\pm$0.001 & 1.37$\pm$0.01 & 1.16$\pm$0.01 \\ \cline{2-6} 
 & RE & 0.98$\pm$0.01 & 0.265$\pm$0.001 & 2.22$\pm$0.01 & 2.1$\pm$0.015 \\ \hline
\multirow{2}{*}{Theorem} & AP & 0.97$\pm$0.01 & 0.219$\pm$0.002 & 1.05$\pm$0.02 & 0.89$\pm$0.02 \\ \cline{2-6} 
 & RE & 0.87$\pm$0.08 & 0.267$\pm$0.002 & 1.96$\pm$0.02 & 1.64$\pm$0.15 \\ \hline
\multirow{2}{*}{Diabetes} & AP & 0.98$\pm$0.01 & 0.217$\pm$0.003 & 0.27$\pm$0.01 & 0.23$\pm$0.01 \\ \cline{2-6} 
 & RE & 0.95$\pm$0.04 & 0.262$\pm$0.001 & 0.36$\pm$0.01 & 0.31$\pm$0.01 \\ \hline
\multirow{2}{*}{Spambase} & AP & 0.93$\pm$0.01 & 0.233$\pm$0.003 & 0.96$\pm$0.02 & 0.79$\pm$0.02 \\ \cline{2-6} 
 & RE & 0.71$\pm$0.2 & 0.273$\pm$0.004 & 2.04$\pm$0.06 & 1.39$\pm$0.4 \\ \hline
\multirow{2}{*}{KDD99} & AP & 0.99$\pm$0.01 & 0.215$\pm$0.001 & 1.06$\pm$0.01 & 0.91$\pm$0.01 \\ \cline{2-6} 
 & RE & 0.93$\pm$0.04 & 0.263$\pm$0.001 & 1.71$\pm$0.01 & 1.53$\pm$0.06 \\ \hline
\multirow{2}{*}{CAPTCHA} & AP & 0.99$\pm$0.01 & 0.215$\pm$0.001 & 0.80$\pm$0.01 & 0.68$\pm$0.01 \\ \cline{2-6} 
 & RE & 0.97$\pm$0.02 & 0.264$\pm$0.001 & 1.22$\pm$0.01 & 1.12$\pm$0.03 \\ \hline
\end{tabular}
\end{table*}

After the exploration phase, 2000 attack samples are generated in the exploitation phase. The Effective Attack Rate (EAR) and diversity metrics are presented in Table~\ref{tbl:real_word_exploit} for both the AP and the RE attacks. It is seen that an EAR of $97.7\%$ in the case of AP and $>91.2\%$ for the RE attacks, is obtained on average. This is seen even though the defender's model is perceived to have a high accuracy as per Table~\ref{tbl:real_world_se}. Accuracy of classifiers is of little significance if the model can be easily evaded. The high effective attack rate for all 10 cases, highlight the vulnerability of classification models and the misleading nature of accuracy, in an adversarial environment, irrespective of the data application domain. The high EAR of the RE attacks, indicate that partial reverse engineering and the linear approximation of the defender's model surface is sufficient to launch an effective attack against it. This can be seen for the KDD99 dataset, which has a reverse engineering accuracy of $55\%$ while its EAR for the RE attack was $93\%$. This is because, generating a high accuracy on the training dataset is not the goal of the RE approach. It is more concerned with generating a large number of diverse attack samples which would be classified as legitimate. This is possible even with partial reverse engineering. While a high RE accuracy indicates a high EAR (consider Cancer dataset), it is not a required condition for the RE attack, making it of practical use in high dimensional spaces.

The diversity of the RE attacks is higher than the AP attacks, on all three metrics, indicating - a larger spread of attacks, lower collocation of points and a uniform distribution in the attack space. This high diversity is obtained for RE, while still maintaining a reasonable high attack rate. The AP attacks, produces lower diversity but has high attack accuracy than the RE attacks. This is because the number of explored anchor points was $>50\%$ (Table~\ref{tbl:real_world_se}), allowing a large scale AP attack to be feasible. The AP attack is therefore an attractive quick attack strategy in high dimensional spaces, irrespective of the attack domain, application type and the model used. The effectiveness of the RE attack depends on the ability of the surrogate model $C'$ to represent the space of \textit{Legitimate}ly classified samples by $C$. The reverse engineering task is dependent on the availability of enough probing budget and the complexity of the boundary represented by $C$. This is the cause for the higher variability in the EAR values for RE attacks in Table~\ref{tbl:real_word_exploit}, as opposed to the AP attacks, where attacks are more tightly packed with the obtained anchor points, leading to lower variability. 

\begin{figure}[t]
  \centering
  \includegraphics[width=0.8\linewidth]{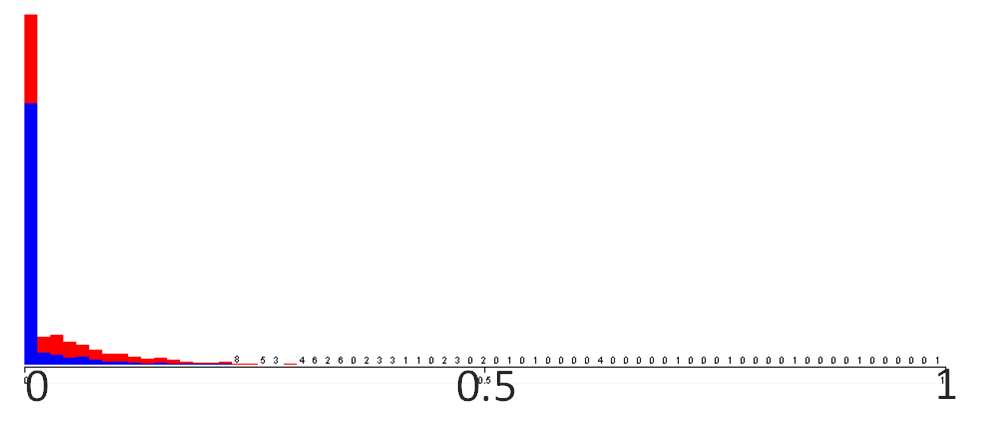}
   \caption{Distribution of Feature \#5 for Spambase dataset, showing a heavy tail. (Red - Malicious, Blue - Legitimate)}
  \label{fig:spam}
\end{figure}

The RE approach's EAR is low for the Credit, the Theorem and the Spambase datasets. In case of the Credit and Theorem dataset, the defender's accuracy is low, indicating a nonlinear separation/ inseparability of samples. The RE accuracy approaches close to the defender's accuracy, but since the original model $C$ has low accuracy, the reverse engineered model can only be so good. For the Spambase dataset, the majority of the features follow a heavy tailed distribution as shown for Feature \#5 in Fig.~\ref{fig:spam}. In such distributions, random sampling in the range [0,1] on each features is not the best choice. Integrating domain information which is commonly known, as in the case of text datasets having heavy tails, can be beneficial. However, following a domain agnostic approach here, a $71\%$ attack rate is still achieved, indicating the viability of such attacks. 

\subsubsection{Experiments with Non-Linear Defender Model}
\label{sec:nonlinear}

The SEE framework considers a black box model for $C$. As such, it is developed as a generic data driven attack strategy, irrespective of the defender's model type, training data or the model parameters. To demonstrate the efficacy of these attacks under a variety of defender environments, experiments with different non linear black box models for $C$ are presented here. Particularly, the following defender models were evaluated: K-Nearest Neighbors classifier with k=3 (kNN) \citep{cover1967nearest}, SVM with an radial basis function kernel with gamma of 0.1 (SVM-RBF) \citep{xiaoyan2003model}, C4.5 Decision Tree (DT)\citep{quinlan1993c4}, and a Random Forest of 50 models (RF)\citep{breiman2001random}, as shown in Table~\ref{tbl:non_linear}. The attacker's model is kept the same as before and the experiments are repeated for each of the defender's model. Average values of EAR over 30 runs are reported in Table~\ref{tbl:non_linear}. 

\begin{table*}[t]
\centering
\caption{Effective Attack Rate (EAR) of AP and RE attacks, with non linear defender's model (Low EAR values are italicized.)}
\label{tbl:non_linear}
\begin{tabular}{l|c|c|c|c|c|c|c|c|}
\cline{2-9}
\multicolumn{1}{c|}{\textbf{}} & \multicolumn{2}{c|}{\textit{\textbf{kNN}}} & \multicolumn{2}{c|}{\textit{\textbf{SVM-RBF}}} & \multicolumn{2}{c|}{\textit{\textbf{DT}}} & \multicolumn{2}{c|}{\textit{\textbf{RF}}} \\ \hline
\multicolumn{1}{|l|}{\textit{\textbf{Dataset}}} & \textit{\textbf{AP}} & \textit{\textbf{RE}} & \textit{\textbf{AP}} & \textit{\textbf{RE}} & \textit{\textbf{AP}} & \textit{\textbf{RE}} & \textit{\textbf{AP}} & \textit{\textbf{RE}} \\ \hline
\multicolumn{1}{|l|}{Digits08} & 0.89 & 0.96 & 0.97 & 0.89 & 0.87 & 0.63 & 0.85 & \textit{0.48} \\ \hline
\multicolumn{1}{|l|}{Credit} & 0.96 & 0.78 & 0.94 & 0.53 & 0.79 & \textit{0.42} & 0.79 & \textit{0.33} \\ \hline
\multicolumn{1}{|l|}{Cancer} & 0.99 & 0.99 & 0.99 & 0.99 & 0.97 & 0.89 & 0.99 & 0.98 \\ \hline
\multicolumn{1}{|l|}{Qsar} & 1 & 0.99 & 0.99 & 0.99 & 0.96 & 0.76 & 0.99 & 0.99 \\ \hline
\multicolumn{1}{|l|}{Sonar} & 0.99 & 0.98 & 1 & 1 & 0.97 & 0.62 & 0.99 & 0.95 \\ \hline
\multicolumn{1}{|l|}{Theorem} & 0.97 & 0.813 & 0.95 & 0.5 & 0.95 & 0.79 & 0.62 & 0.78 \\ \hline
\multicolumn{1}{|l|}{Diabetes} & 0.99 & 0.935 & 0.99 & 0.9 & 0.83 & 0.63 & 0.88 & 0.61 \\ \hline
\multicolumn{1}{|l|}{Spambase} & 0.93 & 0.99 & \textit{0.48} & 0.84 & \textit{0.08} & \textit{0.11} & 0.99 & 0.98 \\ \hline
\multicolumn{1}{|l|}{KDD99} & 0.99 & 0.93 & 1 & 0.99 & 0.89 & 0.54 & 0.92 & \textit{0.27} \\ \hline
\multicolumn{1}{|l|}{Captcha} & 0.99 & 0.92 & 0.99 & 0.92 & 0.97 & 0.83 & 0.93 & 0.89 \\ \hline
\end{tabular}
\end{table*}

The AP approach is minimally affected by the choice of defender's model, with Table~\ref{tbl:non_linear} showing a high EAR for all defender models. The drop in case of Spambase, is attributed to the heavy tailed distributions as explained in Fig.~\ref{fig:spam}. In case of the decision trees, the model trained for Spambase, focuses only on a few key features to perform the classification. Random probing attacks, space out the attack samples across dimensions, without considering their feature importance to classification. This leads to skipping over the key features in the attack generation, making the attacks less effective. However, this could be compensated by performing partial reverse engineering and using a smaller exploitation radius. 

The RE results are significantly dependent on the defender's choice of model. In case of nonlinear data separation, as in the Credit and the Theorem datasets, the linear approximation is a bad choice and this is reflected in the low attack rate. In all other cases, the low attack rate is attributed to the over simplification of the understanding of the models, which in case of the decision tree and random forest tend to be complicated in high dimensional spaces. However, in a majority of the cases it is seen that a $\>$50\% attack rate is still possible with the same linear SVM model used by the adversary. This makes the SEE framework generally applicable to attack classification systems, without explicit assumptions about model types, application domain or the parameters of classification. The efficacy of these approaches, highlights the vulnerability of classifiers to purely data driven attacks, requiring only feature space information.

\subsubsection{Experiments with Google Cloud Prediction Service}
\label{sec:gcp}

To demonstrate the applicability of the RE and the AP techniques on real world remote black box classifiers, we performed experiments using the Google Cloud Prediction API\footnote{\url{https://cloud.google.com/prediction/}}. This API provides machine learning-as-a-service, by allowing users to upload datasets to train models, and then use the trained model to perform prediction on new incoming samples. Google's Prediction API, provides a black box prediction system, as they have not disclosed the model type or the technique used for learning, to the best of our knowledge. As such, this provides for an ideal test of the \textit{SEE}'s attack models, where the defender is remote, accessed from a client and has no information about the defender's models  \citep{papernot2016transferability}. We use the API's Python client library to access the cloud service, and the results on the three cybersecurity datasets are shown in Table~\ref{tbl:gcp}.

The results of the experiment demonstrate that the AP and the RE attacks are effective in attacking the defender's classifier, by generating a high EAR over all datasets. The diversity of the RE approach is seen to be higher for the RE attacks on all three metrics of $\sigma$, $KNN-dist$ and $MST-dist$, indicating the variability of attacks achieved using the RE approach, in a real world setting. Furthermore, the RE accuracy in case of the Spambase dataset (48.1\%) highlights that, linear approximation and partial reverse engineering are sufficient to launch an effective RE attack (EAR=1). These experiments use the same exploration budget ($B_{Explore}$=1000) as the previous sections, to generate attacks of high accuracy and high diversity. In a truly blind-folded setting, where we have no prior information about the defender's classifier, a budget of 1000 ($\approx$ \$0.5)\footnote{\url{https://cloud.google.com/prediction/pricing}} samples indicates the relative ease with which classifiers can be evaded and the need for a more comprehensive defense strategy, beyond a static machine learning model. 

\begin{table*}[t]
\centering
\caption{Results of AP and RE attacks using Google Cloud Prediction API as the defender's black box}
\label{tbl:gcp}
\begin{tabular}{l|c|c|c|c|c|c|}
\cline{2-7}
 & \multicolumn{2}{c|}{Spambase} & \multicolumn{2}{c|}{KDD99} & \multicolumn{2}{c|}{CAPTCHA} \\ \hline
\multicolumn{1}{|l|}{\begin{tabular}[c]{@{}l@{}}Training\\ Accuracy\end{tabular}} & \multicolumn{2}{c|}{93\%} & \multicolumn{2}{c|}{99\%} & \multicolumn{2}{c|}{100\%} \\ \hline
\multicolumn{7}{|c|}{{\ul \textbf{\textit{Attack Metrics}}}} \\ \hline
\multicolumn{1}{|l|}{} & \multicolumn{1}{l|}{\textit{AP}} & \multicolumn{1}{l|}{\textit{RE}} & \multicolumn{1}{l|}{\textit{AP}} & \multicolumn{1}{l|}{\textit{RE}} & \multicolumn{1}{l|}{\textit{AP}} & \multicolumn{1}{l|}{\textit{RE}} \\ \hline
\multicolumn{1}{|l|}{EAR} & 1 & 1 & 1 & 1 & 0.99 & 0.97 \\ \hline
\multicolumn{1}{|l|}{$\sigma$} & 0.216 & 0.264 & 0.218 & 0.265 & 0.218 & 0.265 \\ \hline
\multicolumn{1}{|l|}{KNN-dist} & 1.324 & 2.148 & 1.105 & 1.714 & 0.813 & 1.228 \\ \hline
\multicolumn{1}{|l|}{MST-dist} & 1.127 & 2.078 & 0.944 & 1.645 & 0.695 & 1.131 \\ \hline
\multicolumn{1}{|l|}{\begin{tabular}[c]{@{}l@{}}Accuracy of \\ RE model $C'$\end{tabular}} & \multicolumn{2}{c|}{48.1\%} & \multicolumn{2}{c|}{97.2\%} & \multicolumn{2}{c|}{100\%} \\ \hline
\end{tabular}
\end{table*}

\subsubsection{Effects of Varying $B_{Explore}$ and $R_{Exploit}$}
\label{sec:parameters}

\begin{figure*}[t]
\centering
\subfloat[Effect of $R_{Exploit}$ on EAR]{\includegraphics[width=0.3\linewidth]{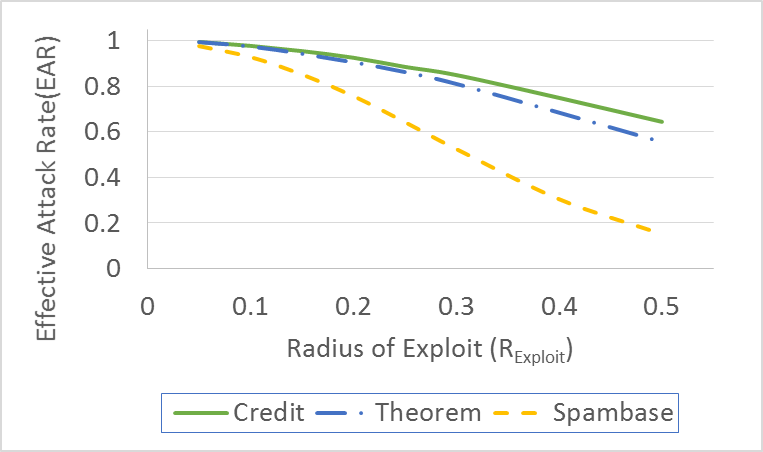}} 
\subfloat[Effect of $R_{Exploit}$ on KNN-dist]{\includegraphics[width=0.3\linewidth]{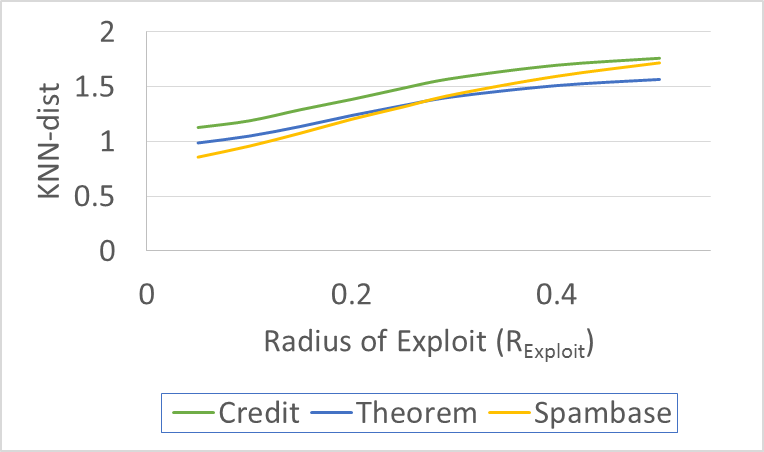}}
\subfloat[Effect of $R_{Exploit}$ on MST-dist]{\includegraphics[width=0.3\linewidth]{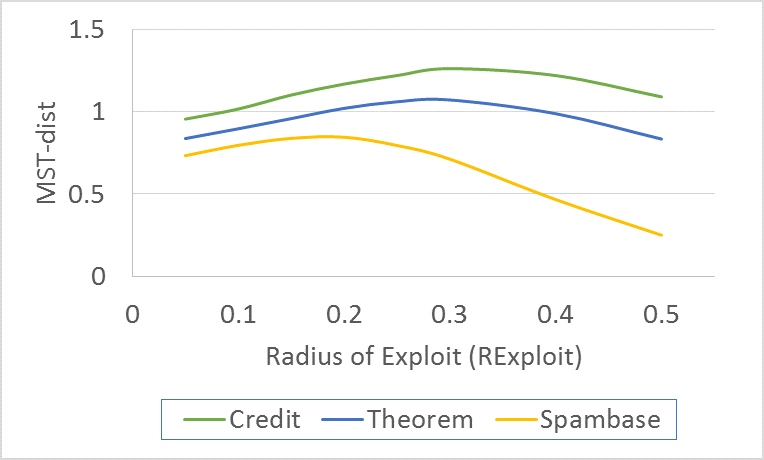}} \\
\subfloat[Effect of $B_{Explore}$ on EAR]{\includegraphics[width=0.3\linewidth]{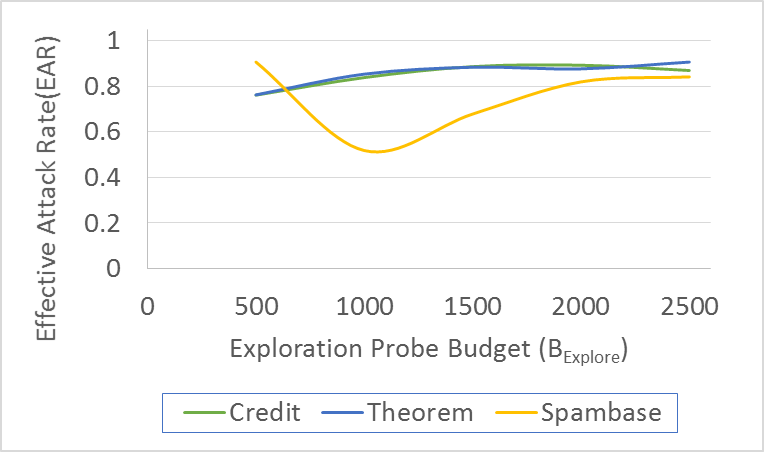}}
\subfloat[Effect of $B_{Explore}$ on KNN-dist]{\includegraphics[width=0.3\linewidth]{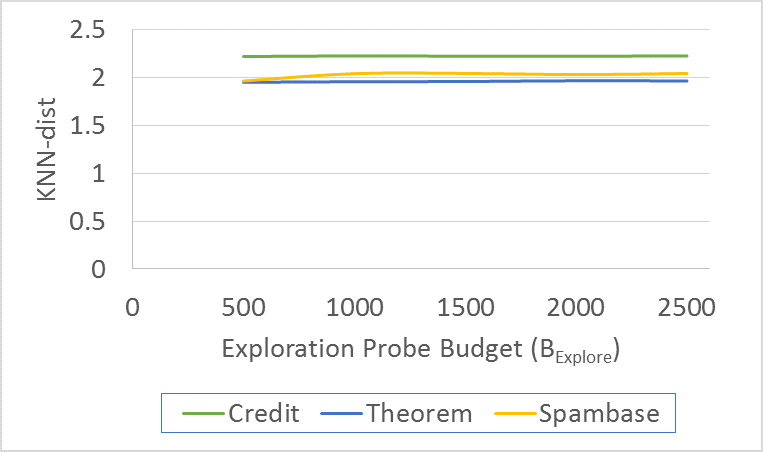}}
\subfloat[Effect of $B_{Explore}$ on MST-dist]{\includegraphics[width=0.3\linewidth]{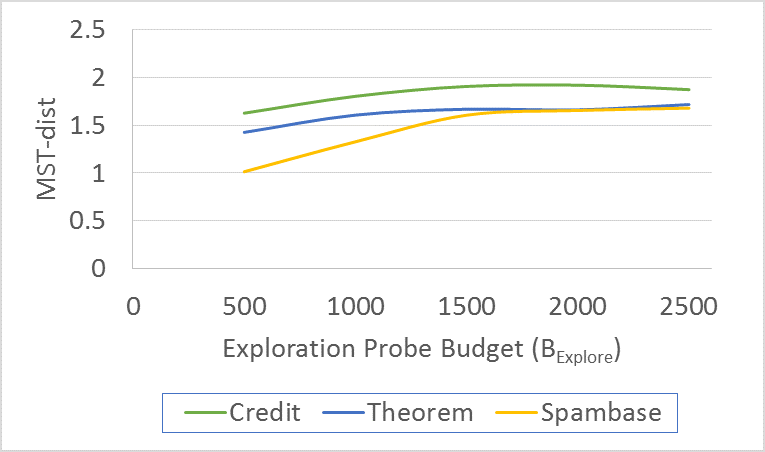}}
 \caption{Effect of changing $R_{Exploit}$, for the AP attacks (\textit{Top}),  and $B_{Explore}$, for the RE attacks (\textit{Bottom}),  on the Effective Attack Rate (EAR) and Diversity (KNN-dist, MST-dist)}
\label{fig:exploit_radius}
\end{figure*}

In evaluating the AP and RE approaches, the $R_{Exploit}$ was kept fixed at 0.1 for AP and 0.5 for RE. This was intuitively motivated, as confidence in attacks would reduce as distance from anchor points increases, as they are the only ground truth information available to the attackers in the AP strategy. Effect of increasing $R_{Exploit}$, on the accuracy and diversity of AP attacks, is shown in Fig.~\ref{fig:exploit_radius}. The Credit, Theorem and the Spambase datasets were chosen for these evaluation, as they have low EAR for the RE approach (Table~\ref{tbl:real_word_exploit}) and could therefore benefit from parameter tuning. Effect of increasing $R_{Exploit}$ to increase diversity of AP attacks, and increasing $B_{Explore}$ to increase  EAR of RE attacks, as viable alternatives to improve performance over these three datasets is analyzed and presented.

The effective attack rate (EAR) reduces with an increase in the exploitation radius, as seen in Fig.~\ref{fig:exploit_radius}a), because attack samples move away from the anchor points. There is an associated increase in the diversity using both KNN-dist and MST-dist measures, as shown in c) and e).  Comparison of diversity and EAR at $R_{Exploit}$=0.5 for the RE and AP approach shows, that for increasing diversity it is much better to switch to the RE approach instead of increasing $R_{Exploit}$ arbitrarily, as the effectiveness of attacks starts dropping rapidly with increased radius. The drop in MST-dist in Fig.~\ref{fig:exploit_radius} e) is due to the reduction of the size of the Effective Attack set (EA).

As increasing diversity for AP approach leads to a drop in EAR, we investigate if we can increase the EAR of the RE approach while maintaining its high diversity.  Increasing the exploration budget increases the EAR, due to availability of labeled training data for the reverse engineered model, leading to better learning of the data space. The increase in EAR ultimately plateaus, as per the Probably Approximate Learning(PAC) principles \citep{haussler1990probably}, indicating that it is not necessary to arbitrarily keep increasing this budget. The knee point is seen in Fig.~\ref{fig:exploit_radius}b) (around 1500 for all datasets). After the knee point, the EAR of all three datasets is $\>$85\%, and adding more probing budget has little impact on the EAR or the diversity(Fig.~\ref{fig:exploit_radius} d, f). It is necessary to have sufficient probing budget to reach this knee point, to allow effective reverse engineering in complex data spaces. This extra effort provides long term benefits as it leads to increased diversity of attacks. RE is suitable for patient adversaries who want to apply data science in breaking the system. In case of a low budget the AP approach is more suitable, but with the RE strategy the assumption is that the adversary wants to spend time to learn the system before attempting an attack

\section{Why diversity is an important consideration in designing attacks?}
\label{sec:diversty}

Throughout the design and evaluation of the SEE framework, diversity of attacks has been considered as an important goal for the adversary. This was intuitively motivated, as diversity ensures that the attacks have enough variability, so as to make its detection and prevention difficult. In this section, we quantify the effects of diversity on the ability to thwart defenses, especially those based on blacklisting of samples. \textit{Blacklists} are ubiquitous in security applications, as an approach to flag and block known malicious samples \citep{kantchelian2013approaches}. Modern blacklists are implemented using approximate matching techniques, such as Locality Sensitive Hashing, which can detect perturbations to existing flagged samples \citep{prakash2010phishnet}. The goal of an attacker is to avoid detection by these blacklists, as they can make a large number of attack samples unusable with a quick filtering step. With high diversity, it is unlikely that blacklisting a few samples will cause the attack campaign to stop. In case of a diverse attack, the defender will have to resort to choosing between maintaining a huge blacklist of samples, or to remodel the machine learning system, both of which are expensive tasks and require time. 

To empirically evaluate the effect of diversity on blacklisting, a synthetic blacklisting experiment is presented, which simulates the effect of approximate matching filters. The blacklist is maintained as a list \textit{BL}, of previously seen attack samples, with an associated approximation factor: $\epsilon$. An attack is detected if a new sample falls within $\epsilon$ distance of any sample in the blacklist \textit{BL}. The entire blacklisting process is simulated as follows: i) the attackers use the SEE framework to generate $N_{Attacks}$ attack samples which are submitted to the defender model $C$, ii) the defender is assumed to gain information over time about these  $N_{Attacks}$ samples and then proceeds to blacklist them by storing them in \textit{BL}, iii) The attacker, still unaware of the blacklisting, continues to use its existing explored information (AP or RE model) to generate additionally more $N_{Attacks\_New}$ attack samples. The effectiveness of the blacklisting process is computed as the number of effective attacks in $N_{Attacks\_New}$, which are detected by \textit{BL}. The \textit{percentage of attacks stopped}, indicates effectiveness of blacklists and consequently the effect of diversity. A small rate would indicate that blacklisting is not effective in stopping such attacks. 

\begin{figure}[t]
\centering
\subfloat[ Percentage of attacks stopped by blacklist]{\includegraphics[width=0.98\linewidth]{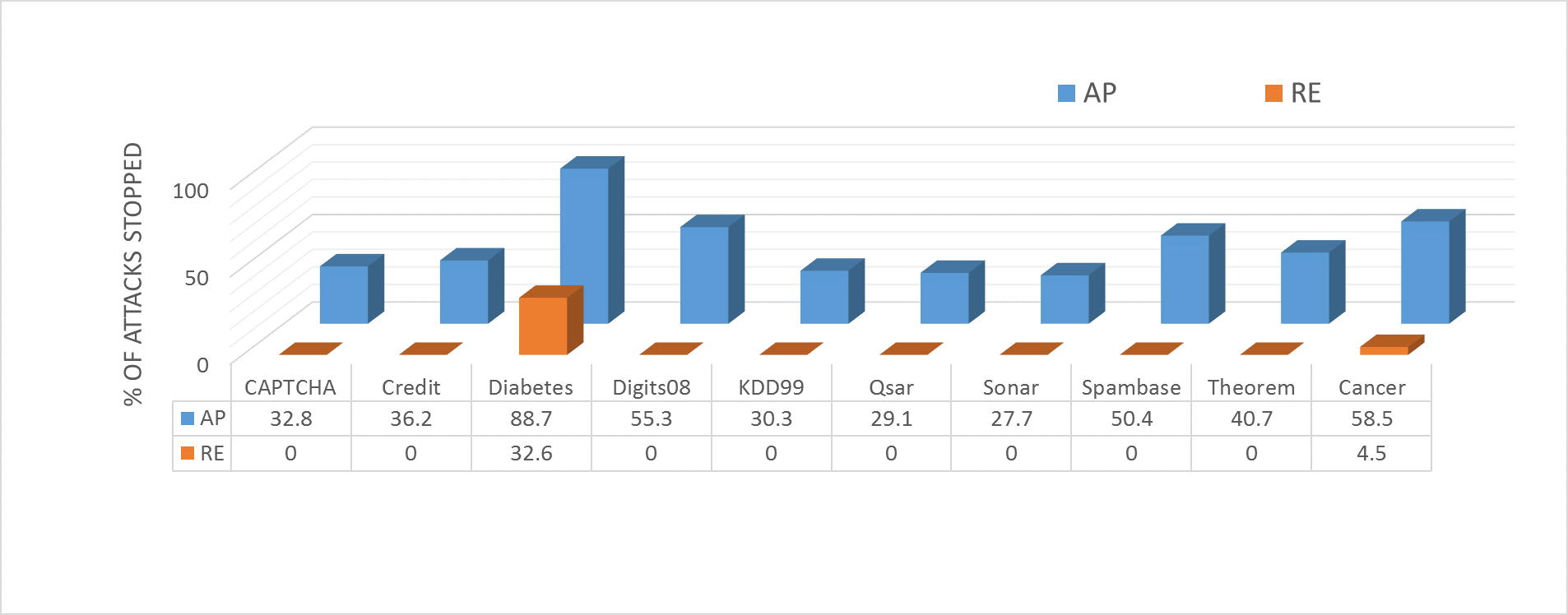}}\\
\subfloat[MST-dist of AP and RE attacks]{\includegraphics[width=0.98\linewidth]{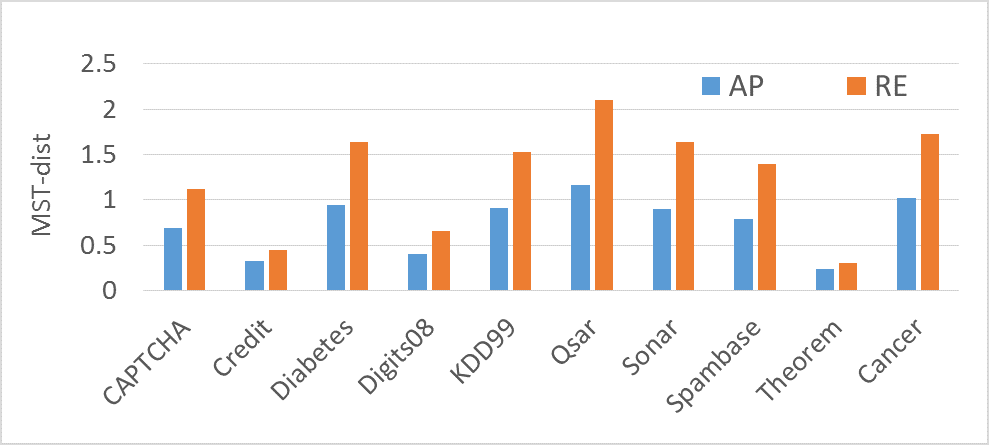}}
\caption{Relationship between diversity and ability to thwart attacks by blacklisting samples.}
\label{fig:diversity}
\end{figure}

Results of the blacklisting experiment, with an approximation factor of $\epsilon=0.1$, on the UCI datasets is shown in Fig.~\ref{fig:diversity}. In order to balance effects of approximation across datasets, the approximation factor is multiplied by $\sqrt[]{d}$, where $d$ is the number of dimensions of the dataset. The $\epsilon$ is chosen so as to balance the efficacy of the blacklists to its false positive rate. Increasing the approximation factor leads to an increase in the number of false positives, which causes legitimate samples to be misclassified as attacks. Limiting false positives is essential, as a blacklist which causes too many false alarms would be impractical. In all experiments, the exploration budget is kept fixed at 1000, the exploitation samples $N_{Attack }$=2000 and an additional 2000 samples($N_{Attack\_New}$) are generated to test the blacklisting effects.

 From Fig.~\ref{fig:diversity}a), it is seen that the AP approach has a higher percentage of stopped attacks than the RE approach, across all datasets. This is a result of the higher diversity of the RE attacks as seen from the MST-dist metric in Fig.~\ref{fig:diversity} b). The high diversity of the RE attack causes blacklisting to be totally ineffective (0 attacks stopped) for 8 out of the 10 datasets. In these cases, increasing the $\epsilon$, as a countermeasure, to stop attacks is not a viable option, due to additional false alarms caused. A higher diversity in RE would force the reevaluation of the security system, leading to redesign, feature engineering and collection of additional labeled samples. All these are time taking and expensive efforts, making the RE approach effective as an attack strategy. As such, if an attacker is sophisticated and has enough probing budget $B_{Explore}$, it can launch a diverse attack campaign, which is harder to stop by adhoc security measures. The AP attack provides for high accuracy and precise attacks, as was seen in Table~\ref{tbl:real_word_exploit}. However, a significant portion of this attack campaign (45\%, on average) is stopped by using a blacklist capable of approximate matching. This experiment aims to highlight the effect of diversity and does not claim to be a concrete defense mechanism. Nevertheless, it intends to be a motivation for further analysis into the efficacy of incorporating such techniques for designing secure machine learning frameworks. Blacklists capable of heuristic matching, could empower classifiers \citep{kantchelian2013approaches} with the ability to recognize perturbed attack samples and as such continue to be effective against anchor points attacks. In case of reverse engineering attacks, a blacklisting based approach was seen to be  ineffective. Taking preemptive counter-measures before deploying the classifier, to successfully detect and mislead attacks, could be a promising direction for dealing with these attacks \citep{hong2016assessing}. 

\section{Conclusion and Future Work}
\label{sec:conclusion}

In this paper, an adversary's view of machine learning based cybersecurity systems is presented. The proposed Seed-Explore-Exploit framework, provides a data driven approach to simulate probing based exploratory attacks on classifiers, at test time. Experimental evaluation on 10 real world datasets shows that, even models having high perceived accuracy ($>$90\%), can be effectively circumvented with a high evasion rate ($>$95\%). These attacks assumed a black box model for the defender's classifier and were carried out agnostic of the type of classifier, its parameters and the training data used. Evaluation results considering 4 different non-linear classifier's and considering the Google Cloud Platform based prediction service, for the defender's black box, demonstrates the innate vulnerability of machine learning in adversarial environments, and the misleading nature of accuracy in providing a false sense of security. Also, the ability to reverse engineer the defender's classifier, and subsequently launch diverse attacks was demonstrated. Attacks with high diversity were shown to be more potent, as they can avoid detection and thwarting, by countermeasures employing blacklists and approximate matching.

The purpose of this work is to make the model designers aware of the nature of attacks that invade classification systems, from a purely data driven perspective. Wearing the white hat, we draw attention to the vulnerabilities introduced by using classifiers in cybersecurity systems. As Sun Tzu says in The Art of War- 'To know your Enemy, you must become your
Enemy'\citep{tzu1963art}. We hope that this work serves as background and motivation for the development and testing of  novel machine learning based security frameworks and metrics. Use of moving target defense strategies \citep{hong2016assessing} and dynamic adversarial drift handling techniques \citep{kantchelian2013approaches, tsethi}, are promising directions warranting further research. Future work will concentrate on developing and analyzing preemptive strategies, in the training phase of the classifiers, to facilitate reliable attack detection and relearning for effective recovery.  




\section*{References}

\bibliographystyle{elsarticle-harv} 
\bibliography{references}




\end{document}